\documentclass[runningheads]{llncs}
\usepackage{graphicx}
\usepackage{subfig}
\newcommand{\etal}{\textit{et al. }}
\begin{document}
\title{KPTransfer: improved performance and faster convergence from keypoint subset-wise domain transfer in human pose estimation}
\titlerunning{KPTransfer: keypoint subset-wise domain transfer in human pose estimation}
%
\author{Kanav Vats \and
Helmut Neher \and
Alexander Wong \and
David A. Clausi \and
John Zelek}
\authorrunning{K. Vats et al.}
%
\institute{Department of Systems Design Engineering, University of Waterloo \\ \email{\{k2vats, hneher, a28wong, dclausi, jzelek\}@uwaterloo.ca}}
\maketitle              
\begin{abstract}

  In this paper, we present a novel approach called KPTransfer for improving modeling performance for keypoint detection deep neural networks via domain transfer between different keypoint subsets.  This approach is motivated by the notion that rich contextual knowledge can be transferred between different keypoint subsets representing separate domains.  In particular, the proposed method takes into account various keypoint subsets/domains by sequentially adding and removing keypoints. Contextual knowledge is transferred between two separate domains via domain transfer. Experiments to demonstrate the efficacy of the proposed KPTransfer approach were performed for the task of human pose estimation on the MPII dataset, with comparisons against random initialization and frozen weight extraction configurations. Experimental results demonstrate the efficacy of performing domain transfer between two different joint subsets resulting in a PCKh improvement of up to 1.1 over random initialization on joints such as wrists and knee in certain joint splits with an overall PCKh improvement of 0.5. Domain transfer from a different set of joints not only results in improved accuracy but also results in faster convergence because of mutual co-adaptations of weights resulting from the contextual knowledge of the pose from a different set of joints.

\end{abstract}

\section{Introduction}

In any keypoint estimation problem, the location of a particular keypoint holds contextual information about the location of another. In pose estimation, for example, the position of the elbows and wrists are naturally constrained, being part of the same limb. Deep keypoint estimation algorithms take advantage of this fact by learning different keypoint locations simultaneously. For instance, deep pose estimation algorithms predict human joint locations together \cite{Cao2017RealtimeM2,mss,Newell2016,Wei2016ConvolutionalPM} or use a two-pipeline framework for body part detection and association \cite{Pishchulin2016DeepCutJS,Tompson:2014:JTC:2968826.2969027}. However, domain transfer between keypoints remains an unexplored area. We hypothesize that domain transfer can be used for utilizing the contextual relationships between keypoint locations for improving convergence and generalization. Since a large number of keypoint detection datasets \cite{coco,andriluka14cvpr,h36m_pami} are available which differ in keypoint location annotations, one obvious question arises: \textit{Can domain transfer between separate keypoint sets help in improving generalization performance and convergence?} \par

\begin{figure*}
	\begin{center}
		\subfloat[Transfer learning]{
		    \includegraphics[width=1\linewidth]{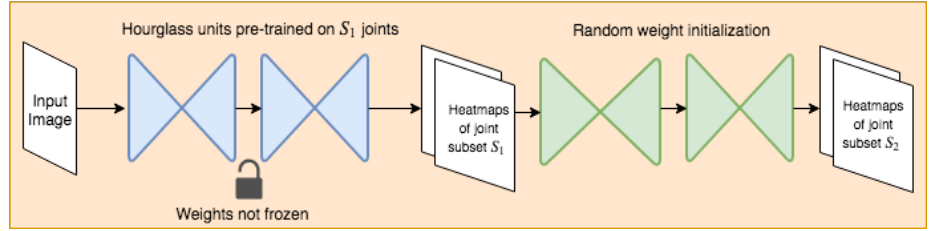}
		}\\
		\subfloat[Frozen weights]{
		    \includegraphics[width=1\linewidth]{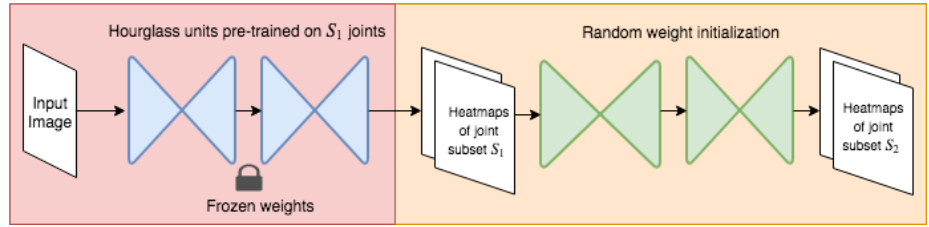}
		}\\ 
			\subfloat[ Random initialization]{
		    \includegraphics[width=1\linewidth]{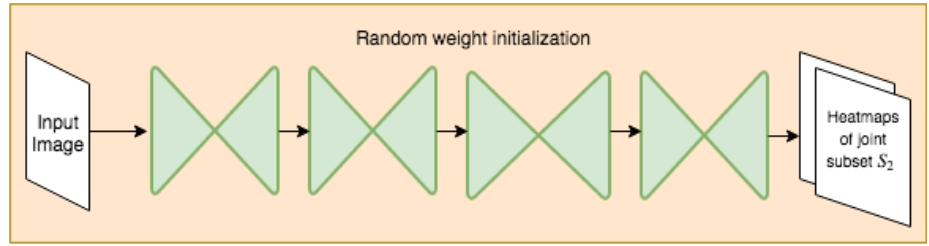}
		}
	\end{center}
	\caption{The figure shows the three experimental configurations used. (a) The transfer learning configuration wherein the two-stack hourglass trained on the subset $S_1$ of joints is used to transfer knowledge to the subset $S_2$ through transfer learning. (b) The frozen weights configuration which is similar to (a), the difference being that the weights of the the two-stack hourglass trained on the subset $S_1$ of joints are frozen. (c) The random initialization configuration where four stacked hourglass units, initialized with random weights are trained on the subset $S_2$ of joints.    }
	\label{fig:configurations}
\end{figure*}

In this paper, we introduce a novel approach termed KPTransfer for performing domain transfer between keypoint subsets representing separate domains. We apply our approach on the problem of 2D human pose estimation. However, our approach is completely task and model agnostic and can be used to evaluate domain transfer using any deep pose estimation model in any other keypoint detection problem such as facial landmark detection and 3D pose estimation. We use the stacked hourglass pose estimation network \cite{Newell2016} to demonstrate that contextual cues can be transferred between different sets of human joints through transfer learning to improve convergence and performance. \par
\begin{figure*}
	\begin{center}
		\subfloat[ This split is done to determine the knowledge transfer between the central body joints and limb joints.]{
		    \includegraphics[width=.5\linewidth,height=.22\linewidth]{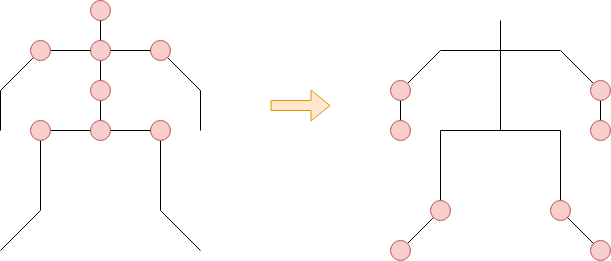}
		}\\
		\subfloat[  Here we include the elbow in our subset $S_1$.]{
		    \includegraphics[width=.5\linewidth,height=.22\linewidth]{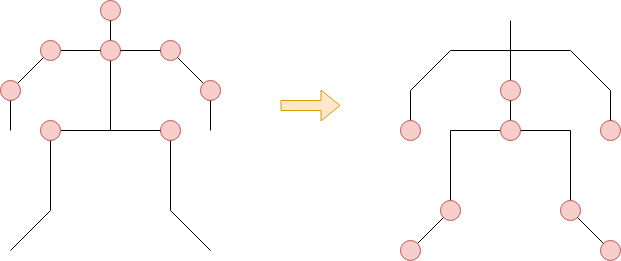}
		}\\ 
			\subfloat[ We include both elbows and knees in our joint subset $S_1$ and determine the accuracy on wrists,hips and ankles in the subset $S_2$.]{
		    \includegraphics[width=.5\linewidth,height=.22\linewidth]{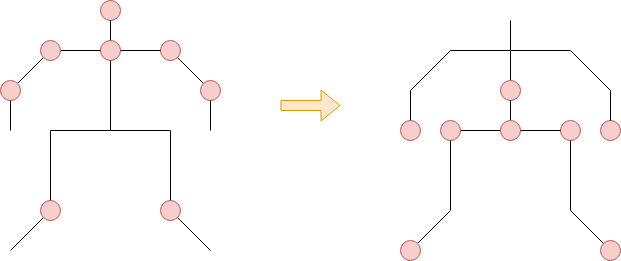}
		}\\ 
			\subfloat[ We now include ankles and wrists in subset $S_1$ and determine the performance on elbows and knees]{
		    \includegraphics[width=.5\linewidth,height=.22\linewidth]{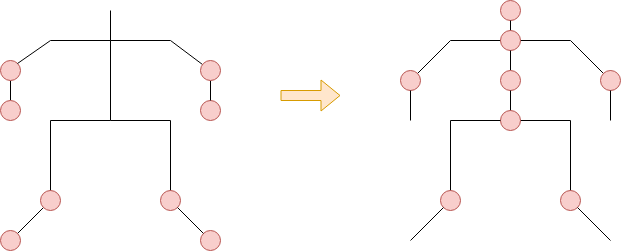}
		}
	\end{center}
	\caption{ The joint subsets $S_1$ and $S_2$ used in the experiments shown as $S_1 \rightarrow S_2$.}
	\label{fig:joint_confs}
\end{figure*}

We perform domain transfer with the help of transfer learning and compare it with frozen weights and random initialization settings. Concretely, for domain transfer, a pose estimation network is first trained on a subset of total joints, after which, a second, bigger network is trained on a different subset of joints, half of which is initialized by the weights of the previous network. Two more settings are investigated: random weight initialization and frozen weights. Random weight initialization is done by initializing \textbf{all} the weights of the second, bigger network randomly. In the frozen weights setting, the second, bigger network is trained after freezing/not updating the weights of the first network. The three settings are illustrated in Fig. \ref{fig:configurations}. We compare the three settings with four different subset splits of human joints (Fig. \ref{fig:joint_confs}) of the  MPII dataset \cite{andriluka14cvpr} and demonstrate that the transfer learning setting results in improved generalization performance and faster convergence. \par

The paper begins by detailing background information related to domain transfer, pose estimation, and the basis for our approach are discussed briefly in Section 2. The Methodology, Section 3, describes the pose estimation network employed for the experiments, the experimental settings and training details. The results and discussion forms Section 4 with Section 5 concluding the paper.

\section{Background}

Domain transfer utilizes information in one domain to help learn tasks in another domain. The two domains involved may represent separate datasets \cite{Chu2016BestPF}, classes in the same dataset \cite{yosinski2014transferable} or different data modalities \cite{Gupta2016CrossMD,hoffman,Zhao2018ThroughWallHP}. Effectiveness of domain transfer techniques has been tested in  various problems like image classification \cite{yosinski2014transferable}, object detection \cite{Hoffman2014LSDALS} and semantic segmentation \cite{Chen2017NoMD}. Transfer learning \cite{yosinski2014transferable} and knowledge distillation \cite{hinton,Zhang2018FastHP} are two popular ways of performing domain transfer. Yosinki \etal \cite{yosinski2014transferable} demonstrate that performing transfer learning by initializing first $n$ layers of a base network with weights learned on approximately half of Imagenet \cite{imagenet_cvpr09} classes with the remaining layers randomly initialized improves generalization performance on the other half of Imagenet classes. Hinton \etal \cite{hinton} introduced knowledge distillation by producing soft probability distribution over targets by modifying the softmax function and introducing an objective function consisting of those soft targets to train a student network. Techniques such as using weight regularizers to make weights of source and target domain networks similar \cite{Rozantsev2018BeyondSW} and adversarial learning \cite{Yang20183DHP,Liu2016CoupledGA} are also employed for domain transfer. \par
With the advent of deep networks, there has been a significant progress in the field of human pose estimation. Toshev \etal \cite{toshev} was among one of the earliest works incorporating deep neural networks (DNN) for pose estimation. Heatmap based pose estimation \cite{Newell2016,Wei2016ConvolutionalPM,Chen2018CascadedPN,Cao2017RealtimeM2} is the most widely used pose estimation technique. In heatmap based methods, joint heatmaps, equal to the number of joint locations present in the images are generated. Each heatmap represents a two dimensional probability distribution where each heatmap pixel represents the probability with which a joint is present in a particular pixel location. Intermediate supervision is commonly used in heatmap based methods, wherein loss is calculated at subsequent stages of the pose estimation network to refine heatmap predictions. Regression based approaches \cite{toshev,Bulat2016HumanPE,Carreira2016HumanPE} are also prevalent in human pose estimation literature, however, their limitation is that the regression function is often sub-optimal. The work presented in this paper makes use of a heatmap based approach \cite{Newell2016}. \par

  Domain transfer between separate keypoints subsets in deep human pose estimation is an unexplored avenue. Multimodal pose transfer methods exist in pose estimation literature. Zhao \etal \cite{Zhao2018ThroughWallHP}  predict human pose from RF signals by transferring visual knowledge from an RGB images based pose estimation model  \cite{Cao2017RealtimeM2} in a multimodal setting. Yang \etal \cite{Yang20183DHP} perform adversarial learning to transfer knowledge between annotated 3D human pose datasets and 2D in-the-wild images. Zhou \etal \cite{Zhou2017Towards3H} use a weakly supervised approach and propose a geometric constrained to regularize depth predictions from 2D in-the-wild images .
Zhang \etal \cite{Zhang2018FastHP} use knowledge distillation \cite{hinton} to transfer knowledge from a teacher pose estimation network to a smaller network. In contrast, we are concerned with the task of domain transfer between separate body keypoint/joint subsets. Human body joints possess information about the location of one another through various constraints imposed by the overall body pose and structure. We present an approach to transfer rich contextual information between human body joints. Our work differs from  approaches such as \cite{yosinski2014transferable} in the way that instead of splitting a dataset based on classes, we have used subsets of human joint locations as separate domains and demonstrate that contextual knowledge can be transferred from one domain to the other using transfer learning. Our approach can be readily extended to other keypoint estimation problems such as facial landmark detection and 3D human pose estimation.

\begin{table}
  \caption{The subsets considered for experiments}
  \label{table:joint_subsets}
  \centering
  \resizebox{\columnwidth}{!}{%
  \begin{tabular}{|ll|l|l|}
    \hline
        &   & Subset $S_1$ & Subset $S_2$ \\
    \hline
     &\textbf{(a)}&   {Head, Neck, Shoulders, Pelvis, Thorax, Hip}  & {Knees, Ankles, Wrists, Elbows}  \\
     &\textbf{(b)}&   {Head, Neck, Shoulders, Elbows, Hip}  & {Knees, Ankles, Wrists, Pelvis, Thorax}  \\
     &\textbf{(c)}&   {Head, Neck, Shoulders, Elbows, Knees}  & {Wrist, Ankles, Hip, Pelvis, Thorax} \\
     &\textbf{(d)}&   {Knees, Ankles, Wrists, Elbows}  & {Head, Neck, Elbows, Knee, Pelvis, Thorax} \\
    \hline
  \end{tabular}%
  }
\end{table}

\section{Methodology}
In this section, the stacked hourglass network, experimental approach and training details are discussed. The MPII \cite{andriluka14cvpr} dataset is used in the experiments which consists of around 28k training images and 11k testing images annotated with 16 body joints. Since the experiments involve evaluation on subsets of 16 annotated joints, a validation set of 3000 images is used for evaluation since the test annotations are not public.

\subsection{Stacked Hourglass Network}
Being the backbone of many state-of-the-art pose estimation algorithms \cite{mss,fppe} on the MPII dataset \cite{andriluka14cvpr}, the stacked hourglass network \cite{Newell2016} is used in our experiments. The hourglass architecture consists of repeated bottom-up and top-down processing in order to utilize features at various scales. Convolution and max pooling layers bring down the input resolution from $64\times64$ pixels to $4\times4$ pixels. After downsampling to a resolution of $4\times4$ pixels, the features are upsampled with nearest neighbour upsampling and are combined with features of the same resolution. Several hourglass units are stacked together such that the output of one hourglass unit serves an input to the next hourglass unit. Intermediate supervision is applied such that mean squared loss is evaluated between the predicted heatmaps and ground truth heatmaps and gradients are back-propagated at every hourglass unit.  The output of the network is a set of heatmaps equal to the number of joints with each pixel in the heatmap representing the probability with which the joint is present at that point.

\subsection{Experiments}


 Domain transfer in keypoint estimation in performed by splitting the dataset joints into two subsets, $S_1$ and $S_2$ containing the same number of joints/keypoints. Both the subsets represent two separate domains. Note that $S_1 \cap S_2$ may or may not be an empty set, such that the two domains differ in atleast one keypoint location. Domain transfer is experimentally determined using three different configurations illustrated in Fig. \ref{fig:configurations}. The network performance is evaluated on the joint subset $S_2$ using the three experiment configurations discussed below:

\subsubsection{Transfer learning}
Since we are interested in determining the domain transfer from domain to another, the two-stack hourglass network trained on the subset $S_1$ of joints is jointly trained in conjugation with another two-stack hourglass network on the subset $S_2$ of joints (Fig. \ref{fig:configurations} (a)). The joint training is done such that supervision is performed for all the four hourglass units.

\subsubsection{Frozen weights}
This configuration is similar to transfer learning except that the weights of the first two-stack hourglass network trained on the subset $S_1$ of joints are frozen. Another two-stack hourglass network is trained on the remaining subset of eight joints $S_2$ such that the features obtained from the frozen network are used as an input for the network trained on subset $S_2$ (Fig. \ref{fig:configurations} (b)). In other words, the loss is calculated only for the last two hourglass units. This is done to avoid the possibility of mutual co-adaptation of weights between the two domains.

\begin{figure*}
	\begin{center}
		\subfloat[ ]{
		    \includegraphics[width=.45\linewidth]{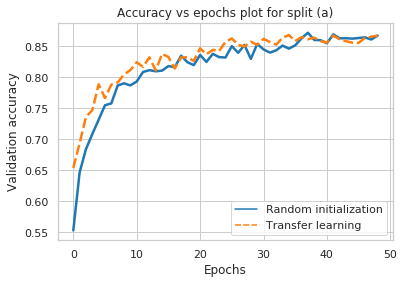}
		}
		\subfloat[  ]{
		    \includegraphics[width=.45\linewidth]{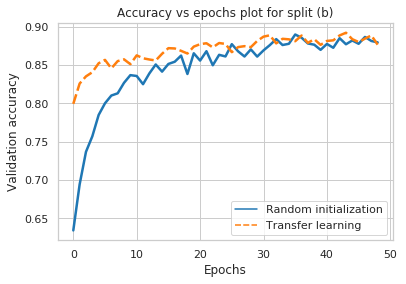}
		}\\
			\subfloat[ ]{
		    \includegraphics[width=.45\linewidth]{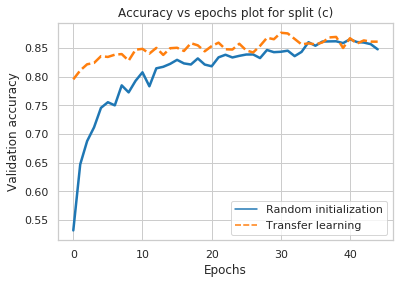}
		} 
			\subfloat[ ]{
		    \includegraphics[width=.45\linewidth]{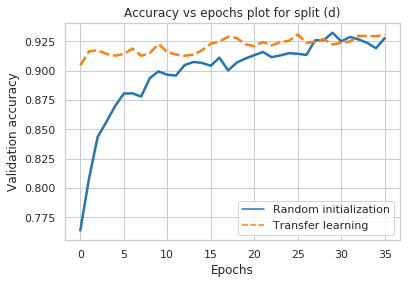}
		}
	\end{center}

	\caption{The figure shows the comparison between the convergence rates of validation accuracy in case of transfer learning and random weight initialization. An important observation is that the convergence and accuracy obtained are complementary to each other such that in split (a) where there is no significant improvement in accuracy, the convergence between random initialization and transfer learning comparable. but, the opposite is observed in split (d), where both convergence and accuracy achieved is better than random initialization. }
	\label{fig:convergence}
\end{figure*}

\subsubsection{Random Initialization}
A stacked hourglass network with four hourglass units with weights initialized randomly is trained on the subset $S_2$ of joints (Fig. \ref{fig:configurations} (c)).

The performance of the three configurations are evaluated with four different splits of subsets $S_1$ of joints and corresponding subset $S_2$ (split (a) -(d)) listed in the Table \ref{table:joint_subsets}. Since, a very large number of subsets are possible, the joint subsets are chosen such that the result of knowledge transfer between adjacent limb joints when compared to random initialization can be evaluated as shown in Fig. \ref{fig:joint_confs}. Note that all the three configurations have roughly the same number of parameters. The goal of the experiments is not to improve the state-of-the-art pose estimation benchmark, but to evaluate keypoint domain transfer.  

\subsection{Training details}

The input image resolution for the network is $256\times256$ pixels and the heatmap resolution is $64\times64$ pixels. For all the experiments, rmsprop optimizer is used. The learning rate is divided by 5 each time the accuracy plateaus. Early stopping is implemented such that the model is said to be converged if there is no improvement in validation accuracy in 10 epochs with each epoch consisting of 8000 iterations. Data augmentation is carried out with $.75 - 1.25$ scale augmentation and $+/- 30$ degree of rotation augmentation. The training is carried on an NVIDIA Geforce Titan X GPU.

\section{Results}

The PCKh metric \cite{andriluka14cvpr} is used to evaluate pose estimation performance. A joint is correctly predicted if the distance between the ground truth and predicted joint location is less than half the length of the head segment. The PCKh values for the joint subset $S_2$ corresponding to the four joint subset splits with respect to the three experiment configurations i.e., transfer learning, frozen weights and random weight initialization is shown in Tables \ref{table:split_a} - \ref{table:split_d} respectively. Note that the \textbf{pelvis and thorax torso joints are not considered in average PCKh computation} since, being at the centre of the body, they are almost perfectly localized in all scenarios and just increase the average values. Figure \ref{fig:convergence} show the validation accuracy vs epochs curves comparisons between random initialization and transfer learning configurations. Apparently, transfer learning from pre-learned weights from other joints not only helps in achieving better accuracy values but also results in much faster convergence when compared to random initialization. The accuracy values shown in the validation accuracy vs epochs plots (Fig. \ref{fig:convergence}) use the PCK metric \cite{Yang:2013:AHD:2554099.2554122}.

\begin{table}
  \caption{PCKh comparison for joint split \textbf{a} }
  \label{table:split_a}
  \centering
  \begin{tabular}{|l|l|l|l|l|l|}
    \hline
      Configuration    & Elbow & Wrist  & Knee  & Ankle & Average \\
    \hline
    Transfer learning & 87.9  & \textbf{84.2}  &  82.9 & \textbf{80.6}  &   83.9 \\
    Frozen weights     &  74.7 & 56  & 72.7  & 66.8 & 67.5 \\
    Random initialization     & 87.9 & 83.9 & \textbf{83.4} & 80.5 &   83.9  \\
    \hline
  \end{tabular}
\end{table}

\begin{table}
  \caption{PCKh comparison for joint split \textbf{b} }
  \label{table:split_b}
  \centering
  \begin{tabular}{|l|l|l|l|l|}
     \hline
      Configuration    & Wrist & Knee  & Ankle  & Average \\
    \hline
    Transfer learning & \textbf{84.1}  & 84.5  &  81.5 & 83.4    \\
    Frozen weights & 69.7     &    73.0    & 62.6  & 68.43  \\
    Random initialization  & 83.6  & \textbf{85}  & \textbf{81.6} & 83.4   \\
     \hline
  \end{tabular}
\end{table}

\begin{table}
  \caption{PCKh comparison for joint split \textbf{c} }
  \label{table:split_c}
  \centering
  \begin{tabular}{|l|l|l|l|l|}
   \hline
      Configuration    & Wrist & Hip  & Ankle  & Average \\
   \hline
    Transfer learning & \textbf{84}  & \textbf{87.1}  &  \textbf{80.1} & \textbf{83.7}    \\
    Frozen weights      &   71.2     & 83.7   & 69.9  & 74.9  \\
    Random initialization  & 82.9  & 87.0  & 79.7 & 83.2   \\
   \hline
  \end{tabular}
\end{table}

\begin{table}
  \caption{PCKh comparison for joint split \textbf{d} }
  \label{table:split_d}
  \centering
  \begin{tabular}{|l|l|l|l|l|}
   \hline
      Configuration    & Head & Elbow  & Knee  & Average \\
    \hline
    Transfer learning & 96.8  & \textbf{88.3} & \textbf{84.4} &  \textbf{89.8}    \\
    Frozen weights  & 91.1   & 87.9       & 82.9   &  87.3  \\
    Random initialization  & \textbf{97.1}  & 87.9  & 83.3 &  89.4  \\
     \hline
  \end{tabular}
\end{table}
\subsection{Discussion}

A number of interesting observations can be made:
\begin{enumerate}
    \item Firstly, it can be observed from the splits (a) - (d) that features transferred from frozen weights of one domain i.e., subset $S_1$ do not achieve good accuracy on the second domain i.e subset $S_2$, when compared to random initialization and transfer learning. This is because, in the frozen weights configuration, since the loss is not computed on the first two hourglass units, there is no mutual co-adaptation of weights between the two domains.
    
    \item From Table \ref{table:split_a}, in the case of split (a), it can be seen that transfer learning from torso joints improves accuracy on the wrists  when compared to random initialization (83.9 vs 84.2), but does not have a considerable impact on the other limb joints. However, as elbows (split (b)) and ankles (split (c)) are included in subset $S_1$, we find that the performance on  joints such as wrist (Table \ref{table:split_b} and \ref{table:split_c}) and ankle (Table \ref{table:split_c}) becomes much better, such that, in split (c) the average PCKh over all joints (83.7) becomes better than random initialization (83.2). The effectiveness of domain transfer is further demonstrated in split (d) where wrists and ankles present in domain $S_1$  provide contextual knowledge to domain $S_2$. From Table \ref{table:split_d}, a higher average PCKh value (89.8) for transfer learning demonstrates the success of keypoint-wise domain transfer. The weights learned from one domain co-adapt with the other domain.  
   
    \item From Fig. \ref{fig:convergence} it is observed that the convergence and performance is complementary; i.e., in case of split (a) where, domain transfer does not result in any significant accuracy improvement, the convergence between random initialization and transfer learning configurations is comparable (Fig. \ref{fig:convergence} a). On the other hand, in case of split (d), where domain transfer performs better than random initialization, the convergence of transfer learning case is much better when compared to random initialization (Fig. \ref{fig:convergence} d). Whereas, in the two other cases the convergence is "between" the two extreme situations of split (a) and (d). This is the result of the mutual co-adaptation of the network weights in learning adjacent limb joints. This shows that the weights learned on the one domain helps in better initialization of the cost function of the other domain which leads to faster convergence.
\end{enumerate}
\section{Conclusion}

This paper introduces the KPTransfer approach for evaluating keypoint subset-wise domain transfer. We demonstrate that knowledge can be transferred between keypoint subsets in pose estimation such that the contextual cues present across domains helps in better generalization and faster convergence. This work also opens the door for cross-dataset domain transfer in keypoint estimation. Future work include: (1) Determining the exact joint subsets/domain between which domain transfer is most effective. (2) Extending this work to other problems like facial landmark detection and 3D human pose estimation. (3) Evaluation of the proposed keypoint domain transfer strategy with other pose estimation networks.

\bibliographystyle{splncs04}
\bibliography{iciar}

\end{document}